\def\year{2019}\relax
\documentclass[letterpaper]{article} 
\usepackage{aaai19}  
\usepackage{times}  
\usepackage{helvet}  
\usepackage{courier}  
\usepackage{url}  
\usepackage{graphicx}  
\usepackage{color}
\usepackage{amsmath}
\usepackage{amssymb}
\usepackage[ruled,noline]{algorithm2e}
\title{Meta-Learning with Hessian-Free Approach in Deep Neural Nets Training}

\author{
Boyu Chen\\ 
Fudan University\\
\texttt{17110180037@fudan.edu.cn} \\
\And
Wenlian Lu \\
Fudan University\\
\texttt{wenlian@fudan.edu.cn}\\
\And
Ernest Fokoue\\
Rochester Institute of Technology\\
\texttt{epfeqa@rit.edu}
}

\begin{document}

    \maketitle

    \begin{abstract}
        Meta-learning is a promising method to achieve efficient training method towards deep neural net and has been attracting increases interests in recent years. But most of the current methods are still not capable to train complex neuron net model with long-time training process. In this paper, a novel second-order meta-optimizer, named Meta-learning with Hessian-Free(MLHF) approach, is proposed based on the Hessian-Free approach. Two recurrent neural networks are established to generate the damping and the precondition matrix of this Hessian-Free framework. A series of techniques to meta-train the MLHF towards stable and reinforce the meta-training of this optimizer, including the gradient calculation of $H$. Numerical experiments on deep convolution neural nets, including CUDA-convnet and ResNet18(v2), with datasets of CIFAR10 and ILSVRC2012, indicate that the MLHF shows good and continuous training performance during the whole long-time training process, i.e., both the rapid-decreasing early stage and the steadily-deceasing later stage, and so is a promising meta-learning framework towards elevating the training efficiency in real-world deep neural nets.
    \end{abstract}

    \section{Introduction}
    Meta-learning, often referred to as \emph{learning-to-learn}, has attracted a steady increase of interest from deep learning researchers in recent years \cite{andrychowicz2016learning,chen2016learning,wichrowska2017learned,li2016learning,li2017learning,ravi2016optimization,wang2016learning,finn2017model}. In contrast to hand-crafted optimizers like Stochastic Gradient Descent (SGD) and related methods like ADAM \cite{kingma2014adam} and RMSprop \cite{tieleman2012lecture}, the methodology of meta-learning essentially revolves around the harnessing of a trained meta-optimizer, typically via recurrent neural networks (RNN), to infer the best descent directions, which are used to train the target neural networks, with the finality of achieving a better learning performance. In statistical machine learning, artificial intelligence and data science, meta-learning is increasingly deemed  a promising learning methodology, by virtue of the widely held belief among researchers and practitioners, that ''meta-trained'' neural networks can ''learn'' much ''more effective'' descent directions than their counterparts trained via hand-crafted methods.

    A meta-learning method is essentially twofold: (i) a well-turned neural network that outputs the ``learned'' heuristic descent direction; and (ii) a decomposition mechanism to share and substantially reduce the number of meta-parameters and thereby enhance its generality, so that the trained meta-optimizer can work for at least one class of neural network learning tasks. The most notable decomposition mechanisms in the past few years include the so-called {\it coordinate-wise} framework \cite{andrychowicz2016learning} and the {\it hierarchical} framework \cite{wichrowska2017learned} developed in the context of Recurrent Neural Networks (RNN). However, it is crucial to note that most current meta-learning methods cannot both (a) remain stable with long-training process on complex target network,  and (b) still be more effective than their hand-crafted counterparts \cite{wichrowska2017learned}. Hence, {\it developing an efficient meta-optimizer along with a good framework that is stable with acceptable computing cost}, remains a major challenge that impedes the practical application of meta-learning methods to training deep neural networks.

    In this paper, we propose a novel second-order meta-optimizer, which utilizes the Hessian-Free method \cite{martens2010deep} as its core framework. Specifically, the contribution and novelty of this paper include:
    \begin{itemize}
        \item Successful and effective adaptation of the well-known Hessian-Free method to the meta-learning paradigm;
        \item Achievement of noteworthy learning improvements in the form of substantial reductions in the learning-to-learn losses of the recurrent neural networks of the meta-optimizer;
        \item Demonstrated evidence of sustained non-vanishing learning progress and improvements for long-time training processes, especially in the context of practical deep neural networks, including CUDA-Convnet \cite{krizhevsky2012cuda} and ResNet18(v2) \cite{he2016deep}.
    \end{itemize}

    \subsection*{Related Works}
    Meta-learning has a long history, indeed almost as long as the development of artificial neural network itself, with the earliest exploration attributed to \citeauthor{schmidhuber1987evolutionary}(\citeyear{schmidhuber1987evolutionary}). Many contributions around the central theme of meta-learning appeared soon after the incipient paper, proposing a wide variety of learning algorithms \cite{sutton1992adapting,naik1992meta,hochreiter1997long}. Around the same time, \citeauthor{bengio1990learning}(\citeyear{bengio1990learning}),\citeauthor{bengio1992optimization}(\citeyear{bengio1992optimization}), \citeauthor{bengio1995search}(\citeyear{bengio1995search}) introduced the idea of learning locally parameterized rules instead of back-propagation.

    In recent years, the framework of coordinate-wise RNN proposed by \cite{andrychowicz2016learning} illuminated a promising orient towards a meta-learned optimizer can be employed to a wide variety of neural network architectures, which inspired the current surge in the development of meta-learning. \citeauthor{andrychowicz2016learning}(\citeyear{andrychowicz2016learning}) also adapted the Broyden-Fletcher-Goldfarb-Shanno (BFGS) algorithm \cite{atkinson2008introduction} with the inverse of Hessian matrix regarded as the memory, and coordinate-wise RNN as the controller of a Neural Turing Machine \cite{graves2014neural}. \citeauthor{li2016learning}(\citeyear{li2016learning}) proposed a similar approach but with training RNN of the meta-optimizer by reinforcement learning. \citeauthor{ravi2016optimization}(\citeyear{ravi2016optimization}) further enriched the method suggested and developed by \citeauthor{andrychowicz2016learning}(\citeyear{andrychowicz2016learning}),  by adapting it to the few short learning tasks. \citeauthor{chen2016learning}(\citeyear{chen2016learning}) used RNN to output the queue point of Bayesian optimization to train the neural network, instead of outputting descent directions. For other meta-learning fields, \citeauthor{finn2017model}(\citeyear{finn2017model}) proposed the Model-Agnostic Meta-Learning method by introducing a new parameter initialization strategy to enhance its generalization performance.

    However, the L2L optimizer in \citeauthor{andrychowicz2016learning}(\citeyear{andrychowicz2016learning})'s work, which use coordinate-wise RNN to output the descent direction directly is argued unable to perform stable and continuous loss descent during the whole training process, especially when generalized to complicated neural networks\cite{wichrowska2017learned}. To conquer this, \citeauthor{wichrowska2017learned}(\citeyear{wichrowska2017learned}) proposed a hierarchical framework to implement this {\it learning-to-learn} idea to train large-scale deep neural networks such as Inception v3 and ResNet v2 on ILSVRC2012 with big datasets that yielded better performances in terms of generalization than its predecessors. However, in comparison to the hand-crafted optimizer, e.g. momentum, on large-scale deep neural networks and dataset, this learning-to-learn paradigm developed still has an ample room to be improved.

    \section{Preliminaries}

    Let $f$ denote a neural network driving by a collection of parameters $w$ such that upon receiving input $x$ from some input space say $\mathcal{X}$, the network delivers $z = f(x, w)$, where $z \in \mathcal{Y}$, for some output space $\mathcal{Y}$.  Let $y \in \mathcal{Y}$ be the true label corresponding to $x \in \mathcal{X}$.  The learning process of a neural network  consists of finding among all possible neural networks $f$, the one that minimizes the expected loss $\mathcal{L}(w) = \mathbb{E}[l(f(X;w),Y)]$, where the loss function $l(\cdot,\cdot)$ is a nonnegative bivariate function defined on $\mathcal{Y} \times \mathcal{Y}$, and used to measure the loss $l(z,y)=l(f(x,w),y)$ incurred from using $z = f(x, w)$ as a predictor of $y$.

    Since we always care the loss on a mini-batch, we still use $l$, $x$, $z$, $y$ to note the mini-batch version of loss, input and label from here and do not use the single sample version any more. Furthermore, for simplicity, we shall from here to use $l(;w)$ in place of the evaluation of $l(f(x, w),y)$, whenever and wherever such a use will be deemed unambiguous.

    \subsection{Natural Gradient}

    Gradient descent as an optimization tool permeates most machine learning processes. The essential goal of the gradient descent method is {\it to provide the direction in the tangent space of the parameter $w$ that decreases the loss function the most}. The well-known {\em first-order gradient} is the fastest direction with respect to the Euclidean  $l_{2}$ metric, and is the basis of most gradient descent algorithms in practice, like Stochastic Gradient Descent (SGD) and  virtually all momentum-driven learning methods like those introduced by \cite{rumelhart1986learning,kingma2014adam,tieleman2012lecture}.

    However, as argued by \citeauthor{amari1998natural}(\citeyear{amari1998natural}), the $l_{2}$ metric of the {\it parameter's tangent space in fact assumes that all the parameters have the same weight in metric but does not take the characteristics of the neural network into consideration}. In addition, this metric does not possess the parameter invariant property \cite{martens2010deep,amari1998natural}. To circumvent this limitation, the concept of {\it natural gradient} for neural networks was developed \cite{park2000adaptive,martens2010deep,martens2014new,desjardins2015natural,martens2015optimizing}. One of a general definition is
    \begin{align*}
        \nabla_{w}^{n}l=\lim_{\epsilon\rightarrow 0} \frac{1}{\epsilon} \mathop{\arg\min}_{d, m(w, w+d)<\frac{\epsilon^2}{2} } (l(;w+d) - l(;w))
    \end{align*}
    where the metric is defined as $m(w,w+d) = l(f(x, w), f(x, w+d))$. Assuming (1). $l(z, z) = 0$, for all $z$; (2). $l(z, z') \ge 0$, for all $z$ and $z'$; (3). $l$ is differentiable with respect to $z$ and $z'$ which is true for the mean square loss and the cross-entropy loss, the metric $m(w,w')$ has the following expansion
    \begin{equation}
        m(w, w+d) = \frac{1}{2} d^{\top} H d + o(\|d\|_2^3),\ H=\frac{\partial z}{\partial w}^{\top}H_l\frac{\partial z}{\partial w}\label{H_define}
    \end{equation}
    where $\frac{\partial z}{\partial w}$ is the Jacobian matrix of $f(x,w)$ with respect to $w$ and $H_l = \frac{\partial^2}{\partial z^2}l(z, z')|_{z=z'=f(x, w)}$ is the Hessian matrix of $l(z,z')$ with respect to $z$ when $z=z'=f(x, w)$. Hence, $H$ is a Generalized Gauss-Newton matrix (GGN) \cite{schraudolph2002fast} and the natural gradient is specified as
    \begin{equation}
        \nabla_{w}^{n}l = \mathop{\arg\min}_{\|d\|_{H}=1} \langle d, \frac{\partial l}{\partial w}\rangle = - \alpha' H^{-1}\frac{\partial l}{\partial w} \label{nature define}
    \end{equation}
    where $\|d\|_{H} = \sqrt{d^{\top} H d}$ and $\alpha'=1/\|H^{-1}\frac{\partial l}{\partial w}\|_{H}$ is the normalization scalar. More specially, if $l(z,z')$ is the cross-entropy loss, then $H$ is the Fisher information matrix, which agrees with the original definition in \cite{amari1998natural}.

    It's been found in many applications that natural gradient performs much better than gradient descent \cite{martens2014new}. However, the calculation of the natural gradient during the learning process for deep neural networks, is fraught with tough difficulties: specifically,  basically calculating $H^{-1}$ directly usually cost unacceptable time in implement, and for deep neural networks, {\it calculating $H$ on a small mini batch of the training data always causes $H$ to lose rank, which leads to the instability of calculating $H^{-1}$.} For second difficulty, one alternative is to use the damping technique \cite{martens2011learning,lecun1998efficient}, which consists of using $\bar{H}$ in place of $H$, with  $\bar{H} = H + \lambda I$, where $\lambda$ is a positive scalar. However, it turns out that selecting the value of $\lambda$ is sensitive: if $\lambda$ is too large, then the natural gradient degenerates to the weighted gradient; on the other hand, if $\lambda$ is too small, the natural gradient could be too aggressive due to the low rank of $H_{l}$ on a mini batch of the training data. One of a well-known auto adaptive damping technique is
Levenberg-Marquardt heuristic \cite{more1978levenberg}, which still needs additional calculation of $l(;w+d)$ periodicity during training.

    \subsection{Natural Gradient by Hessian-Free Approach}\label{HF}
    Due to the above difficulty associate with the high computational cost of $H^{-1}$, it makes sense to avoid any method that needs to directly compute $H^{-1}$. One of the earliest Hessian-Free methods for neural networks was proposed by \cite{martens2010deep,martens2011learning}, and was used to calculate the natural gradient for deep neural networks. The key idea of the Hessian-Free method is twofold: (i) calculate $Hv$; and (ii) calculate $H^{-1}v$.

    First, to achieve $Hv= \frac{\partial z}{\partial w}^{\top}H_l\frac{\partial z}{\partial w}v$ (equation (\ref{H_define})), we can calculate in turn (1). $\mu=\frac{\partial z}{\partial w}v$, (2). $u=H_{l}\mu$, and then (3). $Hv=\frac{\partial z}{\partial w}^{\top}u$. \cite{pearlmutter1994fast,wengert1964simple} suggested a special difference forward process to be used to calculate $\mu=\frac{\partial z}{\partial w}v$; $u=H_{l}\mu$ is easy when $H_{l}$ is of low rank. And, it is notable that $(Hv)^{\top}= u^{\top}\frac{\partial z}{\partial w}$ is a standard backward process. Also, this difference forward and  standard backward processes can be applied to calculate $\bar{H}v=Hv+\lambda v$ as well.

    Second, with the efficient calculation of $Hv$, the natural gradient $H^{-1}v$ can be approximated by the Preconditioned Conjugate Gradient method (PCG) \cite{atkinson2008introduction}, since it only requires the methodology for calculating $H(\cdot)$, and does not need any other information from $H$. This iterative process can be captured as follows:
    \begin{align}
       x_n, r_n = \text{PCG}(v, H, x_0, P, n, \epsilon)\label{PCG}
    \end{align}
    where $x_{0}$ is the initial vector, $P$ is the {\em Preconditioned Matrix}, which takes a positive definite diagonal matrix in practices, $n$ is the number of iterations, and $\epsilon$ is the error threshold for stopping; The output $x_n$ is the numerical solution of equation $Hx=v$ and $r_n$ is the residual vector: $r_n = v - Hx_n$. The more detailed PCG can be viewed in algorithm \ref{PCG} of the supplementary material. It should be highlighted that the choice of $x_0$ and $P$ has a substantial effect on the convergence of the PCG method.

    In the Hessian-Free method to train a neural network, around $10\sim 100$ iterations are generally needed for PCG to guarantee convergence at each training iteration of the neural network \cite{martens2012training}, which leads a much higher computational cost than the first-order gradient methods, which prevents this method popular to be used to train a large-scale deep neural networks on big datasets.

    \subsection{Natural Gradient Method by Factorized Approximation}
      Besides Hessian-Free method, there are other developments of natural gradient by exploring efficient factorized approximations of $H^{-1}$ to reduce the high computational cost of calculating $H^{-1}v$. For instance, TONGA \cite{roux2008topmoumoute} and kfac \cite{martens2015optimizing,grosse2016kronecker} are the ripest natural gradient methods but their powers usually rely on the specific neural networks architectures to make sure the $H^{-1}$ approximate to be factorizable. Hence, despite the remarkable performance achieved on these network architectures, they are not the general methodology for any possible networks.

    \section{Meta-Learning with Hessian-Free approach}

     \begin{figure}[ht]
      \centering
      \includegraphics[width=0.45\textwidth]{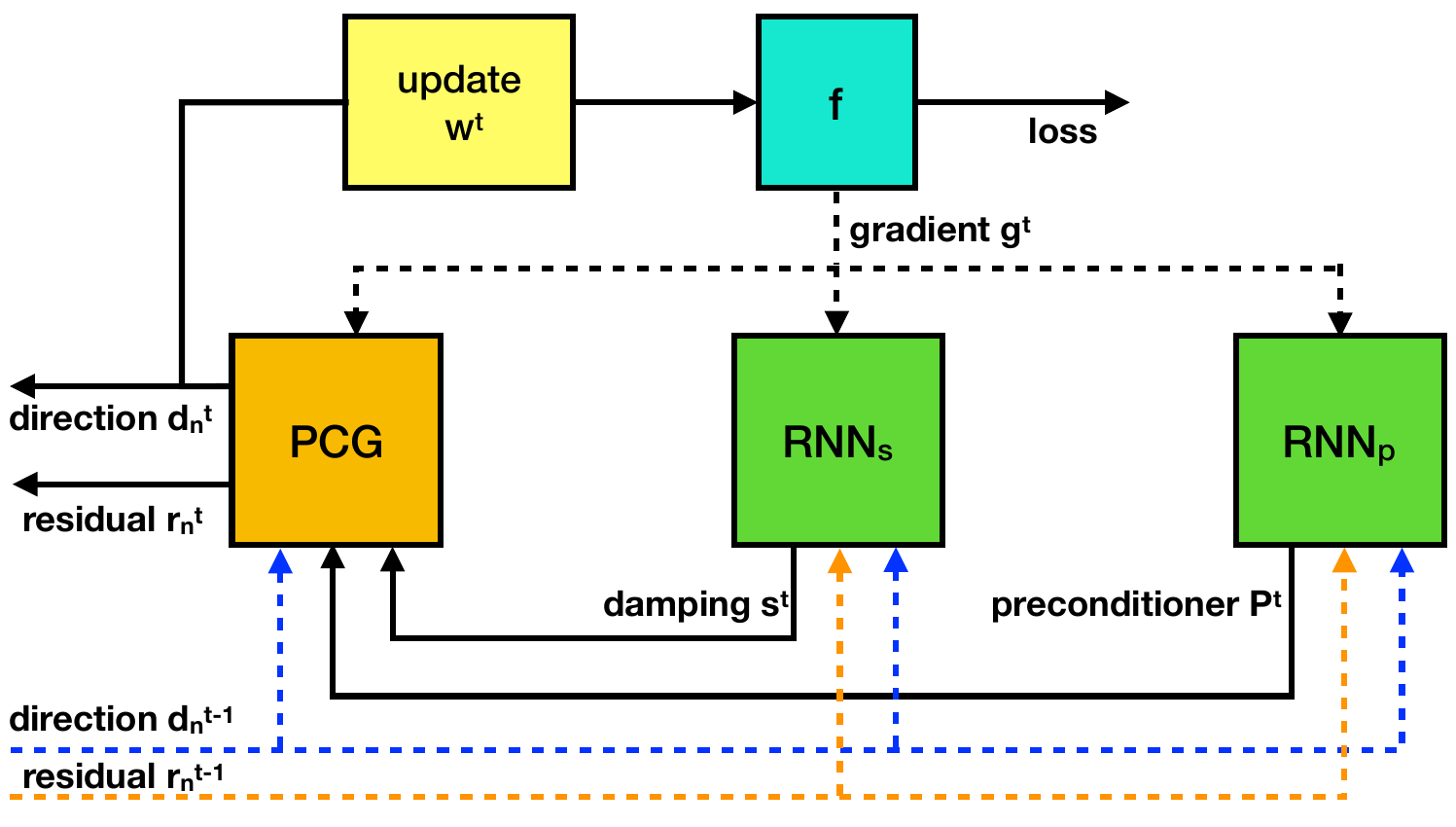}
      \caption{ The architecture of Meta-Learning with Hessian-Free (MLHF) approach: superscript $t$ stands for the step of training the target neural network, and $g$ for the (standard) gradient of $w$, $s$ and $P$ for the damping parameters and Preconditioned matrix respectively. $d_n$ and $r_n$ are the descent direction and the residual generated by MLHF; The dash lines illustrate the directions of meta-trained where the gradient is forbidden to back-propagate (See section \ref{training}).}\label{flowchat}
   \end{figure}

    To conquer the disadvantage of the Hessian-Free approach but still enjoy the advantages of the natural gradient, we propose a novel approach that combines the Hessian-Free method with the meta-learning paradigm. We specifically use a variant of the damping technique with $\bar{H} = H + diag(s)$, where the vector $s=[s_{1},\cdots,s_{n}]\in \mathbb{R}^n$ of parameters has  nonnegative  components, i.e., $s_{i} \ge 0$ for all $i$. The vector $s$ is referred to as the vector of {\em damping parameters}. This variant has a stronger representation capability than the original damping version for which $s=\lambda I$.
    Meanwhile, we generate the damping parameters $s$ and the diagonal preconditioned matrix $P$ by two coordinate-wise RNNs \cite{andrychowicz2016learning}, $\text{RNN}_{s}$ and $\text{RNN}_{p}$ respectively.

    The global computation architecture of this method is illustrated in figure \ref{flowchat} and the specific pseudo-codes can be viewed in Algorithm \ref{ML_with_HF}. With the meta-trained $\text{RNN}_{s}$ and $\text{RNN}_{p}$, at each training step of the neural network $f(x,w)$, $\text{RNN}_{s}$ and $\text{RNN}_{p}$ infer the damping parameter vector $s$ and the   diagonal preconditioned matrix $P$ to the PCG algorithm (\ref{PCG}). The PCG algorithm outputs the approximation of the natural gradient $H^{-1}v$ that gives the descent direction of $l(;w)$.

    \begin{algorithm}
        \DontPrintSemicolon
        \SetKwInOut{Input}{Inputs}
        \SetKwInOut{Output}{Outputs}
        \Input{$n(\le4)$, learning rate $lr$, model $f$, loss function $l$}
        $d_n^{-1} \longleftarrow 0$;\ \ $r_n^{-1} \longleftarrow 0$;\ \ $t \longleftarrow 0$\\
        initialize parameters $w^0$\\
        \While{not terminated}{
        get mini-batch input $x^t$ and label $y^t$\\
        calculate $z^t = f(x^t, w^t)$ and $l^t = l(z^t, y^t)$\\
        calculate gradient $g^t = \frac{\partial l^t}{\partial w^t}$\\
        $d_0^t \longleftarrow d_n^{t-1}$;\ \ $r_0^t \longleftarrow r_n^{t-1}$\\
        $s^t \longleftarrow \text{RNN}_s(d_0^t, r_0^t, g^t)$;\ \ $P^t \longleftarrow diag(\text{RNN}_p(d_0^t, r_0^t, g^t)) $\\
        \textbf{def} $H^tv = \frac{\partial z^t}{\partial w^t}^{\top}H^t_l\frac{\partial z^t}{\partial w^t}v + s^t \odot v, \forall v$ \\
        $d_n^t, r_n^t \longleftarrow \text{PCG}(g^t, H^t, d_0^t, P^t, n, \epsilon=0)$\\
        $w^{t+1} \longleftarrow w^t - lr * d_n^t$\\
        $t \longleftarrow t+1$ \\
        }
        \Output{$w^t$}
        \caption{Meta-Learning with Hessian-Free Approach (MLHF)\label{ML_with_HF}}
    \end{algorithm}

    The network structures of $\text{RNN}_s$ and $\text{RNN}_p$ are coordinate-wise as same as that presented in \cite{andrychowicz2016learning}. In the experiment, we only use $\text{RNN}_s$ and $\text{RNN}_p$ to generate descent directions for the following six type of layer parameters in target network: (i) convolution kernels; (ii) convolution biases; (iii) full connection weights;  (iv) full connection biases; (v) batch-norm  parameter $\gamma$; and (vi) batch-norm parameter $\beta$. The RNNs for each coordinates of the same type layer parameters in the target network share the same meta-parameters, while their state for different coordinates are separated and independent; But for parameter coordinates of different types, RNN meta-parameters are also independent. In addition, the learning rate $lr$ for training the target neural network is fixed to $lr=b_{tr}/b_{mt}$, where $b_{tr}$ is the batch size in target network training and $b_{mt}$ is the batch size in  meta-training, because the magnitude of the damping parameters $s$ work as the learning rate implicitly. On the other hand, the initial vector of PCG at the training iteration $t$ takes the output vector of PCG at the previous iteration $t-1$, as suggested by \cite{martens2012training}.

    \subsection{Training meta-parameters of the $\text{RNN}_s$ and $\text{RNN}_p$}\label{training}

    At each meta-training iteration, we use Back-Propagation-Through-Time (BPTT) \cite{werbos1990backpropagation} to meta-train $\text{RNN}_{s}$ and $\text{RNN}_{p}$ on a short sequence of sampled training process on target network, but with different loss functions as follows. Let $t=1,\cdots,T$ be the iterative count of a sequence of training process on the target network in one meta-training iteration, the loss function of $\text{RNN}_p$ is defined as
    \begin{align*}
        l_p = \frac{1}{T}\sum_t\frac{-\langle d^t_n, g^t\rangle}{\sqrt{\langle d^t_n, H^t d^t_n\rangle}},
    \end{align*}
    where $d^t_n$, $H^t$, $g^t$ are defined in Algorithm \ref{ML_with_HF}.\footnote{Another natural choice is to minimize the square of the norm of $r_n$ in PCG, namely $l_p = \frac{1}{T}\sum_t\|r^t_n\|_2^2$, but it seems not as good as using  (\ref{nature define}), considering the fact that $\|r^t_n\|_2^2$ has quite a different scale, and that it is hard to achieve stability in the initial phase of meta-training.} It can be seen that minimizing $l_{p}$ exactly matches the definition of the natural gradient given in (\ref{nature define}). This is indeed a very encouraging feature as it points to the accuracy of the estimation of the natural gradient by a few iterations of the PCG method. The loss function for $\text{RNN}_{s}$ is defined as
    \begin{eqnarray}
        l_s^t &=& l(f(x^{t+1}, w^{t+1}), y^{t+1}) + l(f(x^{t}, w^{t+1}), y^{t}) \nonumber\\
        && -2\times l(f(x^{t}, w^{t}), y^{t}), \label{ldt}\\
        l_s &=& \frac{\sum_t l_s^te^{l_s^t}}{\sum_t e^{l_s^t}}. \label{ls}
    \end{eqnarray}
    Here $l^{t}_{s}$ is inspired by \cite{andrychowicz2016learning} with some modifications consisting of adding the second item $l(f(x^{t}, w^{t+1}), y^{t})$ in (\ref{ldt}). The motivation for using this term comes from one of the challenges of meta-training, namely that \emph{RNN has the tendency to predict the next input and to fit for it, but the mini-batch $x^t$ is indeed unpredictable in meta-training}, a challenge that tends to cause overfitting,  or make training hard at the early stage. Adding this item in (\ref{ldt}) can reduce such an influence, and thereby stabilize the meta-training process. Thus, $l_{s}$ is the softmax weighted average over all $l^{t}_{s}$. For the sample of the training process on the target network, an experiment replay \cite{mnih2015human,schaul2015prioritized} is also used to store and replay the initial parameters $w^0$.

    \paragraph{Stop gradient propagation}
    During the meta-training $\text{RNN}_{s}$ and $\text{RNN}_{p}$ for predigestion, we do not propagate the gradient of the meta-parameter through $w^t$, $g^t$, $d_0^t$, $r_0^t$ in algorithm \ref{ML_with_HF} and figure \ref{flowchat} in the BPTT rollback, $l(f(x^{t}, w^{t}), y^{t})$ of the third term in (\ref{ldt}), and all $e^{l_s^t}$ in (\ref{ls}).

    Another advantage of stopping back-propagation of gradients of $w^t$, $g^t$, $d_0^t$, $r_0^t$ is to simplify the gradient of multiplication $Hv$ in PCG iterations. In detail, for $u = Hv$ (without the damping part), the $H$'s gradient in the back-propagation progress is not conducted. For the gradient of $v$, we can get $ \frac{\partial l}{\partial v} = H \frac{\partial l}{\partial u}$, which means that the gradient operator of $H(\cdot)$ is itself. By this technique, the calculation of the second-order gradient in meta-training is not necessarily any more, which also reduces GPU memory usage and simplifies the calculation flow graph in practice.

    \subsection{Computation Complexity}

    Compared with the inference of the RNN, the major time consumption of the MLHF method is on the part dedicated to the calculation of the gradient and $Hv$ at each iteration of PCG. As described in section \ref{HF}, calculating $Hv$ mainly involves a special difference forward and a standard backward process. By contrast, calculating the gradient requires a standard forward and backward process. Difference forward is a little faster than the standard forward process, because they can share intermediate results between different iterations of PCG. If one ignores the speed difference between two types of forward process, the time complexity is then found to be $O((n+1)K)$, where $n$ is the maximum number of iterations in PCG, and $K$ is the time that takes to finish once calculation of $Hv$. In the experiments of section \ref{experiment}, we set $n = 4$, which usually results in a training process up to about twice as long as SGD for each iteration.

    \section{Experiments}\label{experiment}

 In this section, we implement the MLHF method of Algorithm \ref{ML_with_HF} via TensorFlow \cite{abadi2016tensorflow}. Specifically, $\text{RNN}_s$ and $\text{RNN}_p$ are two-layered LSTM \cite{hochreiter1997long} with $\tanh(\cdot)$ as the preprocess and a linear map following $\text{softplus}$ as the post-process. Each layer is composed of $4$ units. In the meta-training process, the rollback length of BPTT is set to $10$. We use Adam as the optimizer for the meta-training of the RNNs, and the maximum number $n$ of iterations of PCG is fixed to $4$ by default if there is no other instruction.

In section \ref{CUDA} and \ref{experiments_resnet}, the MLHF performance is evaluated on a simple model (CUDA-convnet) and a more complicated model (ResNet18(v2)) respectively, in comparison with the first-order gradient optimizers, including RMSprop, adam, SGD + momentum (noted as SGD(m)). For CUDA-convnet, we also compare the MLHF with other natural gradient optimizers, including kfac, Hessian-Free with fixed damping (noted as HF(Fixed)), Hessian-Free with the Levenberg-Marquardt heuristic auto-adaptive damping technique (noted as HF(LM)) in section \ref{natural_gradient_comparison}.

We do not compare the MLHF with other natural gradient optimizers on ResNet18(v2), because kfac and HF (Fixed, LM) were reported to prefer a larger batch size ($b=512$) \cite{martens2010deep,grosse2016kronecker} towards stable training, which is out of the limitation of the GPU memory. All the experiments were done on a single Nvidia GTX Titan Xp, and the code can be viewed in \url{https://www.github.com/ozzzp/MLHF}. See table \ref{hyperparameter} in supplementary material for hyper parameters' config of all optimizers.

\subsection{Convnet on CIFAR10}\label{CUDA}
CUDA-Convnet \cite{krizhevsky2012cuda} is a simple CNN with $2$ convolutional layers and $2$ fully connected layers. Here, we use the variant of CUDA-Convnet which drops off the LRN layer and uses the fully connected layer instead of the locally connected layer on the top of the model. This model is simple but has $186k$ parameters, which is still more than the models implemented in the previous learning-to-learn literature. We meta-train a given MLHF optimizer with batch size $b_{mt}=64$ by BPTT on the first $3/5$ training dataset of CIFAR10 \cite{krizhevsky2009learning} for $250$ epochs. After meta-training, we validate this meta-trained optimizer and compare with the first-order optimizers by training the same target model on the remaining $2/5$ training dataset with batch size $b_{tr}=128$. The test performance is inferred on the test dataset.

    \begin{figure}[ht]
        \centering
        \includegraphics[width=0.45\textwidth]{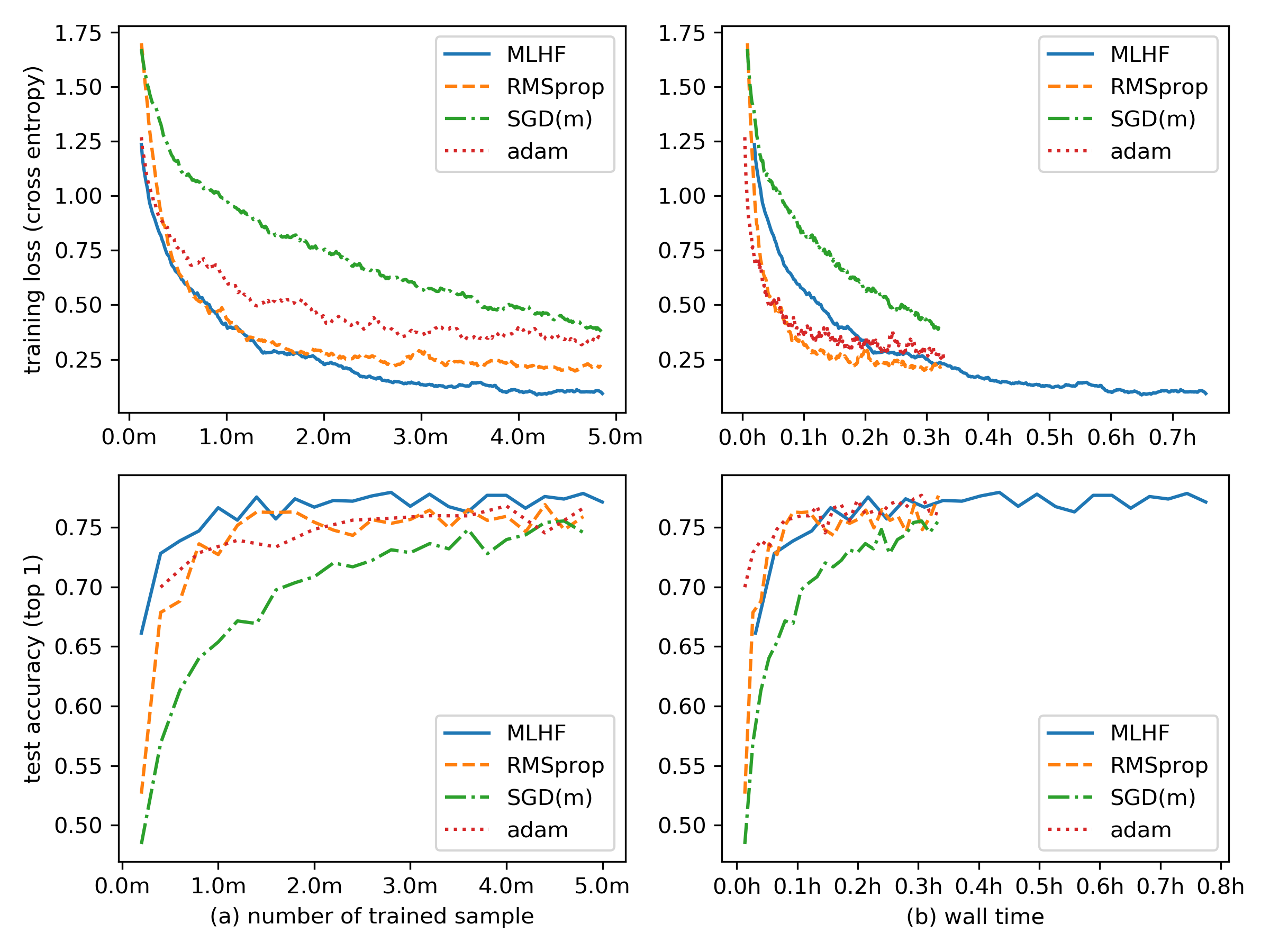}
        \caption{Performance of the training processes of the MLHF compared with the first-order optimizers on the CUDA-Convnet model of the remaining $2/5$ training dataset of CIFAR10 for 250 epochs. Top row : loss (cross entropy) on the training dataset; Bottom row: accuracy on the test dataset; Left column: on the scale of the number of training samples; Right column: on the scale of wall time. Batch size $b_{tr}$ of all optimizers was set to 128.}\label{convnet_on_CIFAR10}
    \end{figure}

Figure \ref{convnet_on_CIFAR10} shows that the MLHF optimizer achieves lower loss in the training data and better inference accuracy in the test data than RMSprop, Adam, and SGD(m) based on same batch size in number of trained sample. However, it is not surprising that the MLHF cost around double amount of time as much as these first-order optimizer per iteration in average.

\subsection{Comparison with Other Natural Gradient Optimizer on Convnet and CIFAR10}\label{natural_gradient_comparison}
    We use the same meta-training configuration as section \ref{CUDA} and compare its performance against other natural gradient based optimizer, including kfac,  HF(Fixed) and HF(LM), except that the batch size $b_{tr}$ is set to 512 for stabilizing these natural gradient optimizer. Specifically, for HF(Fixed) and HF(LM), PCG is run for sufficient iterations (see table \ref{hyperparameter} in supplementary materials) to convergence. In comparison, our MLHF still takes $4$ iterations for PCG, which is far away from convergence (See section \ref{ablation_sec} for details).

    \begin{figure}[ht]
        \centering
         \includegraphics[width=0.45\textwidth]{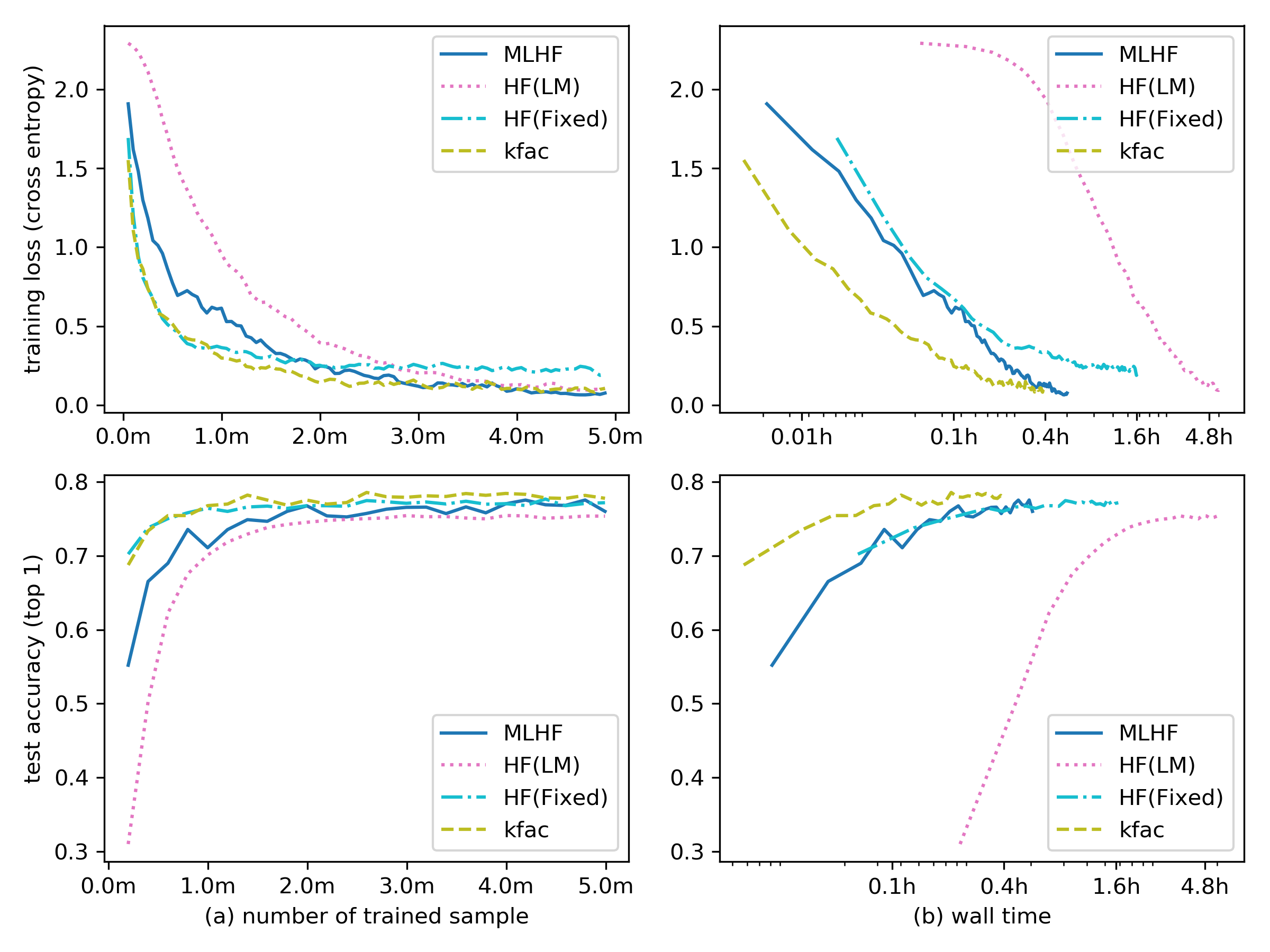}
               \caption{Performance of the training processes of MLHF compared with other natural gradient based optimizers on the CUDA-Convnet model of the remaining $2/5$ training dataset of CIFAR10 for 250 epochs. Top row: loss (cross entropy) on the training dataset; Bottom row: accuracy on the test dataset; Left column: on the linear scale of the number of training sample; Right column: on the logarithmic scale of the wall time. Batch size $b_{tr}$ of all optimizers was set to 512.}
            \label{damping}
        \end{figure}

    As shown in figure \ref{damping}, HF(LM), kfac and MLHF achieve almost the same final loss descent, but the HF(Fixed) has a litter higher final loss descent than the others, although the HF(Fixed) descent rapidly in early stage. In comparison, the HF(LM) descends much flatter during early stage than the others, and obtains the worst generalization performance among all. Compared with the HF(LM) and HF(Fixed), the MLHF performs well on both final loss descent and generalization accuracy during the entire training process. Due to the limited PCG iteration count of the MLHF, it is faster than the  HF(LM) and HF(Fixed) on the scale of the wall time. This evidences that the damping by $\text{RNN}_s$ works well in comparison to the other damping techniques, and the introduce of learning to learn technique is indeed speed up training and get a better performance as well.

The kfac achieves the best performance among all optimizers. One interpretation is that the kfac was doing a lot of online estimation of the approximation of Hessian inverse \cite{martens2015optimizing}. However, this online estimation strongly depends on  handcrafted factorized approximation of Hessian inverse, specified towards given network architecture \cite{martens2015optimizing}. Hence, its Hessian inverse is essentially different from and more stable than those methods based on only a single batch, such as MLHF, HF(Fixed) and HF(LM), which are instead general frameworks and avoid manual designing this factorization approximation.

\subsection{Ablation of $\text{RNN}_p$}\label{ablation_sec}
This subsection aims to verify the efficiency of $\text{RNN}_p$ towards calculating the natural gradient. We use the same  meta -training configuration as section \ref{CUDA} but use the whole training dataset and conduct meta-training the MLHF by the following four configurations:
    \begin{enumerate}
        \item Remove $\text{RNN}_p$ and set iteration count of PCG to $20$.
        \item Remove $\text{RNN}_p$ and keep 4 iteration count of PCG.
        \item Keep $\text{RNN}_p$, but set iteration count of PCG to $2$.
        \item Keep all default.
    \end{enumerate}
We highlight that config (1) has the best performance of PCG but a the largest computation cost.

    \begin{figure}[ht]
        \centering
        \includegraphics[width=0.45\textwidth]{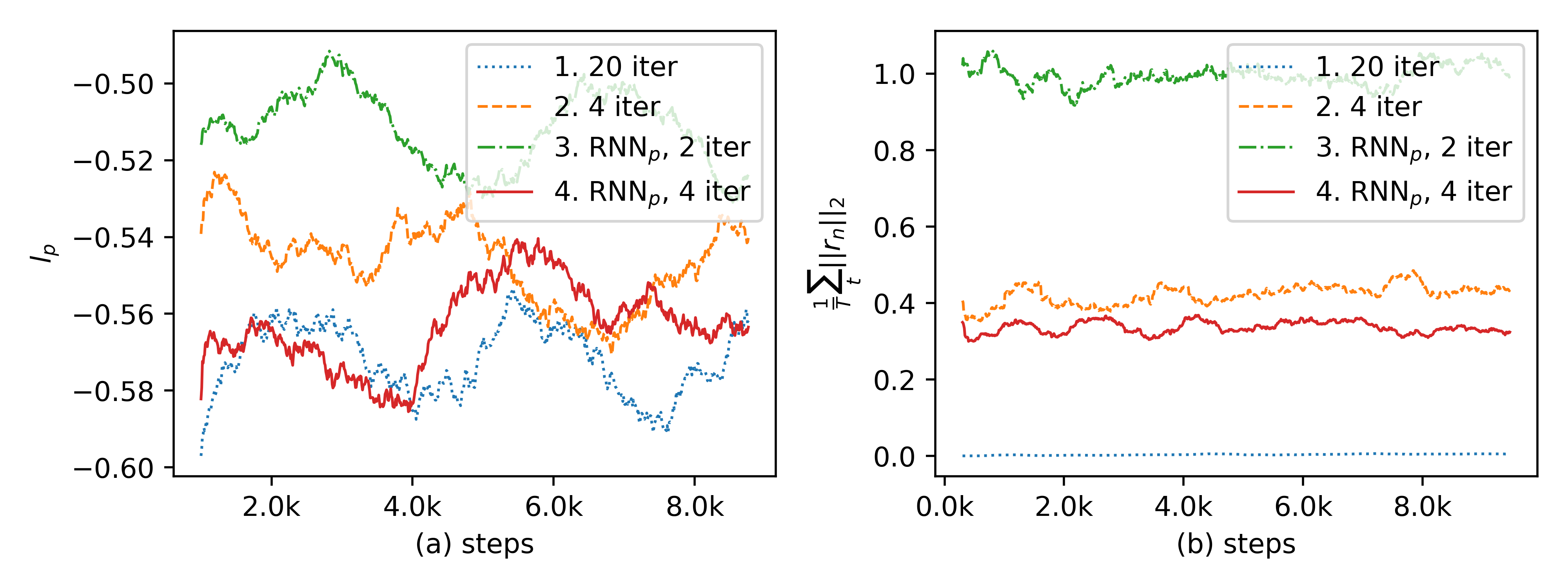}
        \caption{Ablation contrast of $l_p$ (a) and $\frac{1}{T}\sum_t\|r_n\|_2$ (b) in meta-training with respect to iteration step for the underlying four configurations of the MLHF.}\label{Ablation}
    \end{figure}

  Figure \ref{Ablation} (a) and (b) illustrate the following observations. First, with the help of $\text{RNN}_{p}$,  very few ($4$) iterations of PCG (config 4) can estimate the natural gradient as precisely/accurately as a far greater number of iterations of PCG (config 1) measured by $l_{p}$ (Figure \ref{Ablation} (a)); however, 4 iterations is far away from convergence of PCG, in contrast, $20$ iterations (config 1) can guarantee a good convergence of PCG, measured by the mean of $\|r_{n}\|^{2}$ (Figure \ref{Ablation} (b)). Second, in contrast, without $\text{RNN}_{p}$,  few iterations of PCG (config 2) results in a bad estimation of natural gradient and of course far away from convergence of PCG. Finally, we highlight that $4$ iterations could be the optimal number for PCG with the help of $\text{RNN}_{p}$, because further reduction of the number of iteration, i.e., 2 iterations of PCG (config 3), results in both a bad approximation of natural gradient and a bad convergence of PCG.

    \subsection{ResNet on ILSVRC2012}\label{experiments_resnet}
To validate the generalization of the MLHF between different datasets and different (but similar) neural network architectures, we implement a \emph{mini version} of the ResNet \cite{he2016deep} model on whole CIFAR10 training dataset for $250$ epochs, which has $9$ res-block with channel $[16, 16, 16, 32, 32, 32, 64, 64, 64]$, for meta-training. Then we use the meta-trained MLHF to train the ResNet18(v2) on ILSVRC2012 \cite{deng2012ilsvrc} dataset. In the meta-training, the batch size $b_{mt}=128$, while in target training on ILSVRC2012, the batch size $b_{tr}=64$, due to the limitation of GPU memory.

    \begin{figure}[ht]
        \centering
        \includegraphics[width=0.45\textwidth]{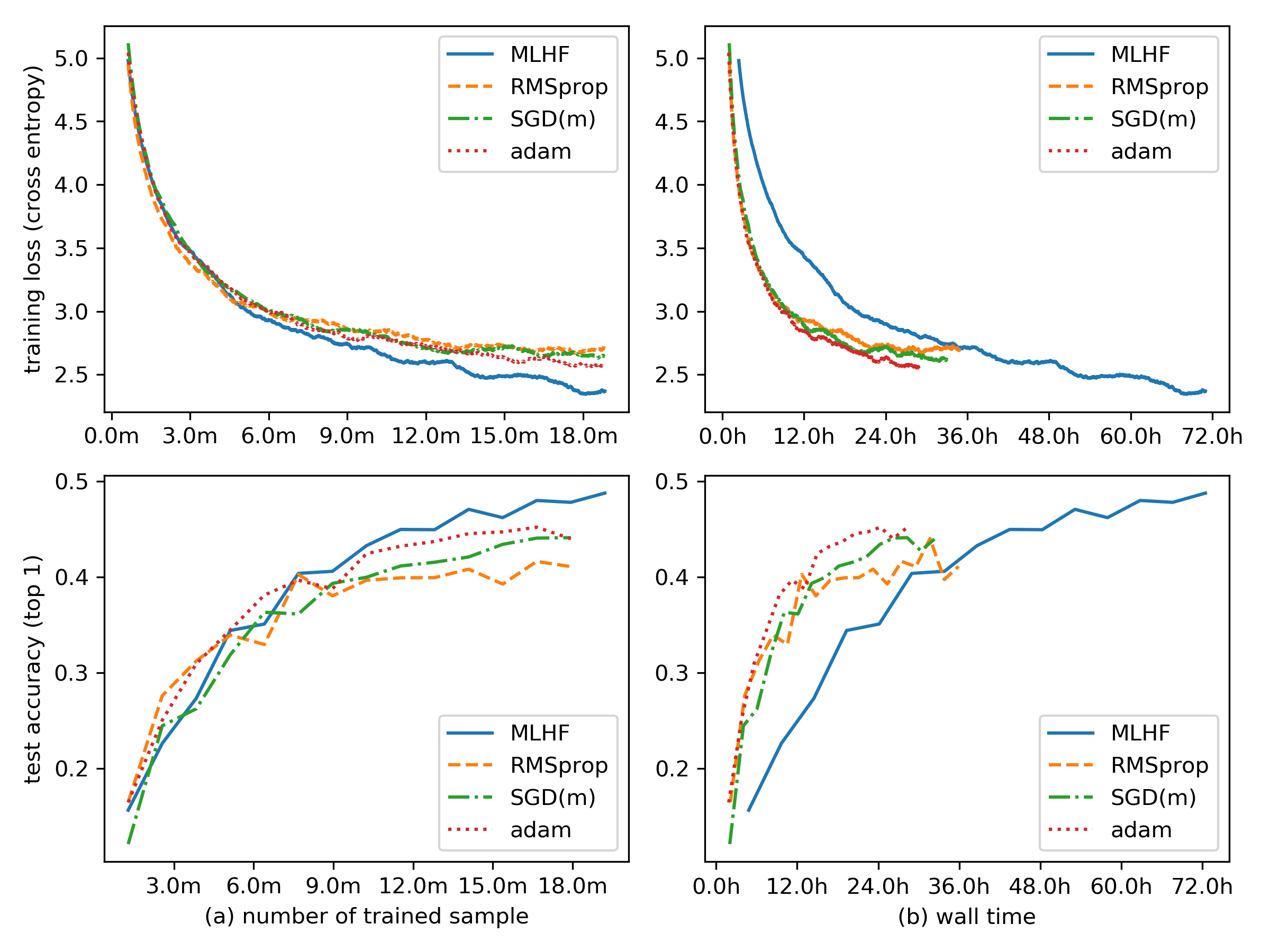}
        \caption{Performance of the training processes of MLHF with other optimizers on ResNet18 (v2) on the dataset ILSVRC2012. Top row: loss (cross entropy) on the training dataset; Bottom row: accuracy on the test dataset; Left column: on the scale of the number of training sample; Right column: on the scale of wall time. Batch size $b_{tr}$ of all optimizers was set to 64.}\label{resnet_on_imagenet}
    \end{figure}

Figure \ref{resnet_on_imagenet} shows that the MLHF achieves the best performance in both training loss and testing accuracy among all evaluated optimizers on the scale of the training sample number. It has also been seen that the MLHF has effective descent progress of the loss function during the whole long-time training, which overcomes the major shortcoming of the previous meta-learning methods \cite{wichrowska2017learned}. However, figure \ref{resnet_on_imagenet} (b) indicates that MLHF costs around double time as much as the first-order optimizer cost per iteration in average.

    \section{Conclusions and Discussions}
In this paper, we proposed and implemented a novel second-order meta-optimizer based on the Hessian-Free approach. We used the PCG algorithm to approximate the natural gradient as the optimal descent direction for neural network training. Thanks to the coordinate-wise framework, we designed two recurrent networks: $\text{RNN}_s$ and $\text{RNN}_p$, to infer the damping parameters and the preconditioned matrix, with a very small number of iterations, the PCG algorithm can achieve a good approximation of the natural gradient with an acceptable computational cost. Furthermore, we used a few specifically designed techniques to efficiently meta-train the proposed MLHF. We have illustrated that our proposed meta-optimizer efficiently makes progress during both the early and later stages of the whole long-time training process in a large-scale neural network with big datasets, and specifically demonstrated this strength of our method on the CUDA-convnet on CIFAR10 and ResNet18 (v2) on ILSVRC2012. The presented meta-optimizer can be a promising meta-learning framework to generalize its performance from simple model and small dataset to large but similar model and big dataset, and elevate the training efficiency in practical deep neural networks.

We present some interpretation of the advantages of the MLHF approach as follows.

\paragraph{Advantage of RNN Damping}
One good choice of damping on each batch is to make $\bar{H}$ approximate to the Hessian Matrix on the whole training dataset. It is speculated that using RNN damping implicitly induces the capability of the method to learn to memorize the history and decode out the diagonal part of the Hessian during meta-training. However, numerical validation to this point is difficult because as far as we know, even on a batch, there is still no effective way to calculate the diagonal part of the Generalized Gauss-Newton matrix.

\paragraph{Stability of MLHF} Compared with L2L, the explanation of the stability of the MLHF is twofold: First, it can be seen that regardless of how $\text{RNN}_s$ is trained, if $\text{RNN}_p$ works well in the sense that $d_n$ approaches near $(H + diag(s))^{-1}g$ well, $ \langle d_n, g\rangle \simeq g^{\top} (H + diag(s))^{-1} g$  is always equal or greater than $0$. This result implies that even if $\text{RNN}_s$ is over-fitting, the loss $l(;w)$ can still decrease, because $d_{n}$ partially follows the standard gradient. Therefore, the training process inherently has a built-in mechanism to efficiently descend gradually even at the early stage. Second, since each coordinate of $d_n$ is determined by all coordinates of $s$ and $P$, it  may result in a good error-tolerance.

Despite the promising performances described above, we are keenly aware of the main limitation of our
proposed method, namely the still relatively high computational cost (even surely much better than the previous Hessian-Free approach), compared with the first-order gradient method. It appears that the price we paid for algorithmic stability is indeed an increase in computational cost.

For the future work, we will evaluate the generalization performance of the MLHF on a more extensive variety of neural networks, including RNN, and RCNN \cite{girshick2015fast}. We also plan to develop the distributed version of the MLHF in order to implement on a larger popular network like ResNet50. Another one of our future orients is to accelerate this MLHF method. We are fully confident, based on our very promising results and performances,  that we can make the learning-to-learn approach exhibit its inherent promised efficacy in the training and effective use of deep neural networks.

    \bibliography{papers}
    \bibliographystyle{aaai}
    
\clearpage
 
\begin{center}
\section*{Supplemental Material}
\end{center}

\setcounter{table}{0}
\setcounter{section}{0}
\setcounter{algocf}{0}
\setcounter{equation}{0}

\section{Preconditioned Conjugate Gradient}
The Preconditioned Conjugate Gradient method (PCG) \cite{atkinson2008introduction}, which captured in the main text as:
    \begin{align}
       x_n, r_n = \text{PCG}(v, H, x_0, P, n, \epsilon)\label{PCG}
    \end{align}
    can be described more detail as in algorithm \ref{PCG}. It is easily to confirm that PCG only requires the methodology for calculating $H(\cdot)$, and does not need any other information from $H$.
        \begin{algorithm}
        \DontPrintSemicolon
        \SetKwInOut{Input}{Inputs}
        \SetKwInOut{aim}{Aim}
        \SetKwInOut{Output}{Outputs}
        \aim{compute $A^{-1}b$}
        \Input{$b$, $A$, initial value  $x_0$,\\ Preconditioned Matrix $P$,\\ maximum iteration number $n$, \\
        error threshold $\epsilon$}
        $r_0 \longleftarrow b - Ax_0$\\
        $y_0 \longleftarrow \textbf{solution of } Py=r_0$\\
        $p_0 \longleftarrow y_0 $;\ \ $i \longleftarrow 0$\\
        \While{$\|r_i\|_2 \ge \epsilon $ and $i \le n$}{
        $\alpha_i \longleftarrow \frac{r_i^{\top}y_i}{p_i^{\top}Ap_i}$\\
        $x_{i+1} \longleftarrow x_i + \alpha_i p_i$;\ \ $r_{i+1} \longleftarrow r_i - \alpha_i Ap_i$\\
        $y_{i+1} \longleftarrow \textbf{solution of }Py=r_{i+1}$\\
        $\beta_{i+1} \longleftarrow \frac{r_{i+1}^{\top}y_{i+1}}{r_i^{\top}y_i}$\\
        $p_{i+1} \longleftarrow y_{i+1} +\beta_{i+1}p_i$\\
        $i \longleftarrow i+1$
        }
        \Output{$x_n\text{ with } x_n\simeq A^{-1}b, \text{residual error } r_i$}
        \caption{Preconditioned conjugate gradient algorithm (PCG)\label{PCG}}
    \end{algorithm}

    the back-propagation via PCG is not complicated as much as common imagination, since the self-gradient property of operator $H(\cdot)$, see section 3.1 for detail in main text.

 \section{Hyper-Parameter Selections in Experiments}
 Table \ref{hyperparameter} detailed note the config of all hyper-Parameters in experiments.
       \begin{table}
      \centering
      \begin{tabular}{c|cccc}
      \textbf{Optimizer} & \textbf{Parameter} & \textbf{Search Range}\\
      \hline
      SGD(m) & $lr$ &  $\{0.1, 0.01, 0.001\}\times b_{tr}/b_{bl}$\\
      & momentum & $\{0.9, 0.99, 0.999\}$ \\
      \hline
      RMSprop & $lr$ & $\{0.1, 0.01, 0.001\}\times b_{tr}/b_{bl}$\\
      & decay & $\{0.9, 0.99, 0.999\}$ \\
      \hline
      Adam & $lr$ & $\{0.1, 0.01, 0.001\}\times b_{tr}/b_{bl}$\\
      & $\beta_1$ & $\{0.9, 0.99, 0.999\}$ \\
      & $\beta_2$ & $\{0.9, 0.99, 0.999\}$ \\
      \hline
      kfac & $lr$ & $\{1, 0.1, 0.01, 0.001\}$ \\
      \emph{(tf official} & damping & $\{1, 0.1, 0.01, 0.001\}$ \\
       \emph{implement)} & cov\_ema\_decay & $\{0.9, 0.99, 0.999\}$ \\
      \hline
      HF(Fixed) & $lr$ & $\{1, 0.1, 0.01, 0.001\}$ \\
      \emph{(our} & $n$ in PCG & 20\\
      \emph{implement)} & $\epsilon$ in PCG & 1e-5 \\
      & $P$ in PCG & $I$\\
      & damping & $\{1, 0.1, 0.01, 0.001\}$\\
      & momentum & 0 \\
      \hline
      HF(LM) & $lr$ & $\{1, 0.1, 0.01, 0.001\}$ \\
      \emph{(our} & $n$ in PCG & 20\\
      \emph{implement)} & $\epsilon$ in PCG & 1e-5 \\
      & $P$ in PCG & $I$\\
      & init damping & $\{1, 0.1, 0.01, 0.001\}$\\
      & decay in LM & $\{2/3, 0.9, 0.99, 0.999\}$\\
      & momentum & 0 \\
      \hline
      MLHF & $lr$ & $b_{tr}/b_{mt}$\\
      & $n$ in PCG & 4\\
      \hline
      & other & default
      \end{tabular}
      \caption{hyper parameters' config of various optimizers in section 4 in the main text. The parameters were chosen from the search range via gird search. Of each experiment, $b_{tr}$ is the batch size of target network training, while $b_{mt}$ is which of meta-training. $b_{bl}$ is the baseline batch size, and sets to 128 for section 4.1 and 256 for section 4.4 in main text.}\label{hyperparameter}
   \end{table}
 
\end{document}